\documentclass{article}
\usepackage{ijcai21}

\usepackage{subfigure}
\usepackage{times}
\usepackage{soul}
\usepackage{url}
\usepackage[hidelinks]{hyperref}
\usepackage[utf8]{inputenc}
\usepackage[small]{caption}
\usepackage{graphicx}
\usepackage{amsmath}
\usepackage{amsthm}
\usepackage{booktabs}
\usepackage{algorithm}
\usepackage{algorithmic}
\usepackage{multirow}
\usepackage{xcolor}
\title{Time Series Data Augmentation for Deep Learning: A Survey}

\author{
    Qingsong Wen\textsuperscript{\rm 1}, Liang Sun\textsuperscript{\rm 1}, Fan Yang\textsuperscript{\rm 2}, Xiaomin Song\textsuperscript{\rm 1}, Jingkun Gao\textsuperscript{\rm 3}\footnote{The work was done when Jingkun Gao was at Alibaba Group.}, Xue Wang\textsuperscript{\rm 1}, Huan Xu\textsuperscript{\rm 2} 
    \affiliations
    \textsuperscript{\rm 1}DAMO Academy, Alibaba Group, Bellevue, WA, USA\\
    \textsuperscript{\rm 2}Alibaba Group, Hangzhou, China\\
    \textsuperscript{\rm 3}Twitter, Seattle, WA, USA
    \emails{
    \{qingsong.wen, liang.sun, fanyang.yf, xiaomin.song, xue.w, huan.xu\}@alibaba-inc.com}, jingkung@twitter.com
}

\begin{document}

\maketitle

\begin{abstract}
Deep learning performs remarkably well on many time series analysis tasks recently. The superior performance of deep neural networks relies heavily on a large number of training data to avoid overfitting. However, the labeled data of many real-world time series applications may be limited such as classification in medical time series and anomaly detection in AIOps. As an effective way to enhance the size and quality of the training data, data augmentation is crucial to the successful application of deep learning models on time series data. In this paper, we systematically review different data augmentation methods for time series. We propose a taxonomy for the reviewed methods, and then provide a structured review for these methods by highlighting their strengths and limitations. We also empirically compare different data augmentation methods for different tasks including time series classification, anomaly detection, and forecasting. Finally, we discuss and highlight five future directions to provide useful research guidance.


\end{abstract}

\vspace{-0.1cm}
\section{Introduction}
\vspace{-0.1cm}

Deep learning has achieved remarkable success in many fields, including computer vision (CV), natural language processing (NLP), and speech processing, etc. Recently, it is increasingly embraced for solving time series related tasks, including time series classification~\cite{IsmailFawaz2019}, time series forecasting~\cite{Han2019}, and time series anomaly detection~\cite{gamboa2017deep}.  
The success of deep learning relies heavily on a large number of training data to avoid overfitting. Unfortunately, many time series tasks do not have enough labeled data. As an effective tool to enhance the size and quality of the training data, data augmentation is crucial to the successful application of deep learning models. The basic idea of data augmentation is to generate synthetic dataset covering unexplored input space while maintaining correct labels. Data augmentation has shown its effectiveness in many applications, such as AlexNet~\cite{krizhevsky2012imagenet} for ImageNet classification.



However, less attention has been paid to find better data augmentation methods specifically for time series data. Here we highlight some challenges arising from data augmentation methods for time series data. Firstly, the intrinsic properties of time series data are not fully utilized in current data augmentation methods. One unique property of time series data is the so-called temporal dependency. Unlike image data, the time series data can be transformed into the frequency and time-frequency domains and effective data augmentation methods can be designed and implemented in the transformed domain. This becomes more complicated when we model multivariate time series where we need to consider the potentially complex dynamics of these variables across time. Thus, simply applying those data augmentation methods from image and speech processing may not result in valid synthetic data. Secondly, the data augmentation methods are also task dependent. For example, the data augmentation methods applicable for time series classification may not be valid for time series anomaly detection. In addition, data augmentation becomes more crucial in many time series classification problems where class imbalance is often observed. In this case, how to effective generate a large number of synthetic data with labels with less samples remains a challenge. 



Unlike data augmentation for CV~\cite{Shorten2019} or speech~\cite{Cui2015}, data augmentation for time series has not yet been comprehensively and systematically reviewed to the best of our knowledge. One work closely related to ours is~\cite{iwana2020empirical} which presents a survey of existing data augmentation methods for time series classification. However, it does not review the data augmentation methods for other common tasks like time series forecasting~\cite{bandara2020improving,hu2020datsing,lee2020stock} and anomaly detection~\cite{lim2018doping,zhou2019beatgan,jingkun20_TAD}. Furthermore, the potential avenues for future research opportunities of time series data augmentations are also missing.  





\begin{figure}[!t]
\centering
    \includegraphics[width=1\linewidth]{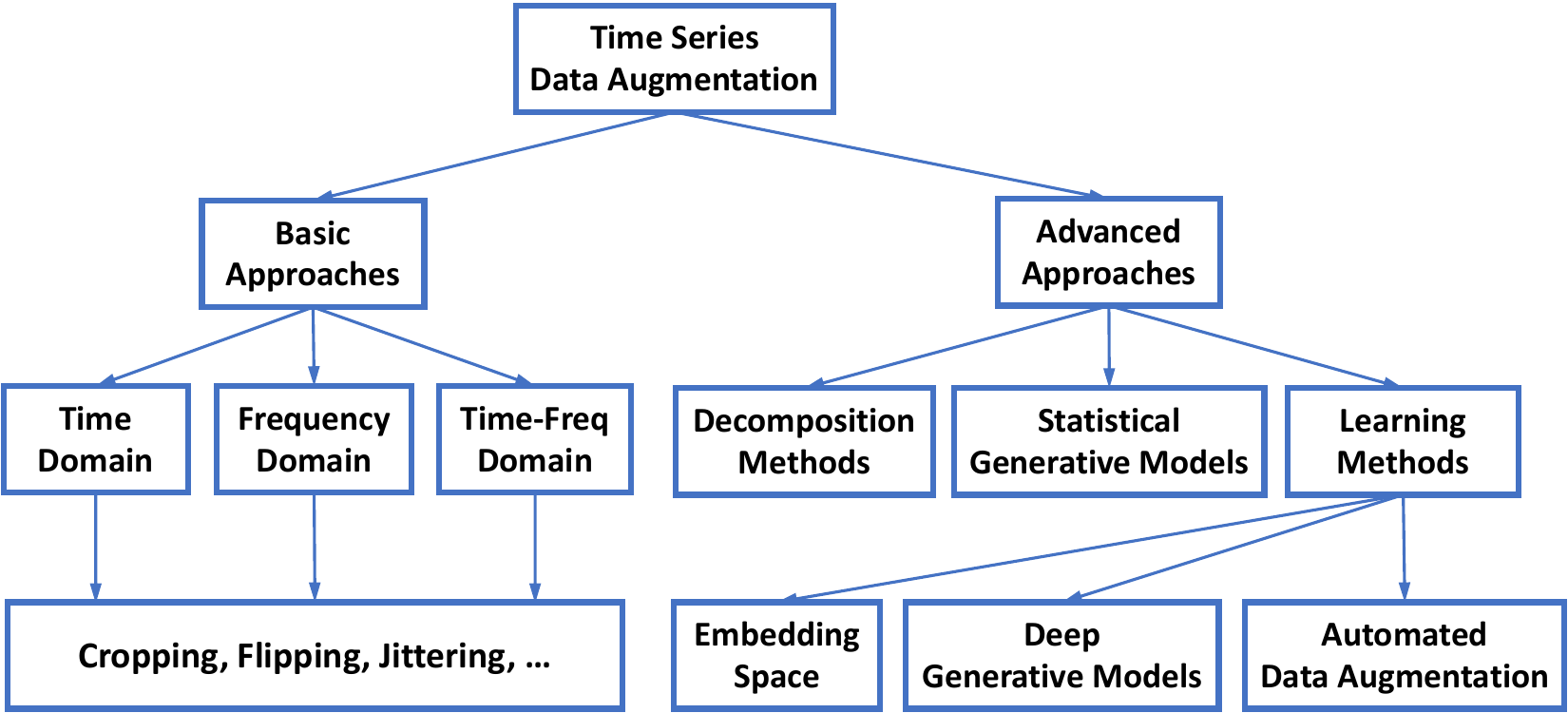}
    \vspace{-0.3cm}
    \caption{A taxonomy of time series data augmentation techniques.}
    \vspace{-0.3cm}
\label{fig:taxonomy}
\end{figure}

In this paper, we aim to fill the aforementioned gaps by summarizing existing time series data augmentation methods in common tasks, including time series forecasting, anomaly detection, classification, as well as providing insightful future directions.
To this end, we propose a taxonomy of data augmentation methods for time series, as illustrated in Fig.~\ref{fig:taxonomy}. Based on the taxonomy, we review these data augmentation methods systematically. We start the discussion from the simple transformations in time domain first. And then we discuss more transformations on time series in the transformed frequency and time-frequency domains. 
Besides the transformations in different domains for time series, we also summarize more advanced methods, including decomposition-based methods, model-based methods, and learning-based methods.
For learning-based methods, we further divide them into embedding space, deep generative models (DGMs), and automated data augmentation methods.
To demonstrate effectiveness of data augmentation, we conduct preliminary evaluation of augmentation methods in three typical time series tasks, including time series classification, anomaly detection, and forecasting.
Finally, we discuss and highlight five future directions: augmentation in time-frequency domain, augmentation for imbalanced class, augmentation selection and combination, augmentation with Gaussian processes, and augmentation with deep generative models.



\vspace{-0.1cm}





\vspace{-0.1cm}
\section{Basic Data Augmentation Methods}
\vspace{-0.1cm}

\subsection{Time Domain}
\label{time-domain-aug}


The transforms in the time domain are the most straightforward data augmentation methods for time series data. Most of them manipulate the original input time series directly, like injecting Gaussian noise or more complicated noise patterns such as spike, step-like trend, and slope-like trend. Besides this straightforward methods, we will also discuss a particular data augmentation method for time series anomaly detection, i.e., label expansion in the time domain. 



{Window cropping} or slicing has been mentioned in ~\cite{le2016data}. Introduced in ~\cite{cui2016multi}, window cropping is similar to cropping in CV area. It is a sub-sample method to randomly extract continuous slices from the original time series. The length of the slice is a tunable parameter. For classification problem, the labels of sliced samples are the same as the original time series. During test time, each slice from a test time series is classified using majority voting. For anomaly detection problem, the anomaly label will be sliced along with value series.

{Window warping} is a unique augmentation method for time series. Similar to dynamic time warping (DTW), this method selects a random time range, then compresses (down sample) or extends (up sample) it, while keeps other time range unchanged. {Window warping} would change the total length of the original time series, so it should be conducted along with window cropping for deep learning models. This method contains the normal down sampling which takes down sample through the whole length of the original time series.

{Flipping} is another method that generates the new sequence $x^{'}_1, \cdots, x^{'}_N$ by flipping the sign of original time series $x_1, \cdots, x_N$, where $x^{'}_t = -x_t$. The labels are still the same, for both anomaly detection and classification, assuming that we have symmetry between up and down directions.

Another interesting perturbation and also ensemble based method is introduced in ~\cite{fawaz2018data}. This method generates new time series with DTW and then ensembles them by a weighted version of the Barycentric Averaging (DBA) algorithm. It shows improvement of classification in some of the UCR datasets.

{Noise injection} is a method by injecting small amount of noise/outlier into time series without changing the corresponding labels. This includes injecting Gaussian noise, spike, step-like trend, and slope-like trend, etc. For spike, we can randomly pick index and direction, randomly assign magnitude but bounded by multiples of standard deviation of the original time series. For step-like trend, it is the cumulative summation of the spikes from left index to right index. The slope-like trend is adding a linear trend into the original time series. These schemes are mostly mentioned in~\cite{wen2019time}

In time series anomaly detection, the anomalies generally last long enough during a continuous span so that the start and end points are sometimes ``blurry". As a result, a data point close to a labeled anomaly in terms of both time distance and value distance is very likely to be an anomaly. In this case, the {label expansion} method is proposed to change those data points and their labels as anomalies (by assign it an anomaly score or switch its label), which brings performance improvement for time series anomaly detection as shown in~\cite{jingkun20_TAD}.





\subsection{Frequency Domain}
\label{freq-domain-aug}

While most of the existing data augmentation methods focus on time domain, only a few studies investigate data augmentation from frequency domain perspective for time series. 

A recent work in~\cite{jingkun20_TAD} proposes to utilize perturbations in both amplitude spectrum and phase spectrum in frequency domain for data augmentation in time series anomaly detection by convolutional neural network. Specifically, for the input time series $x_1, \cdots, x_N$, its frequency spectrum $F(\omega_k)$ through Fourier transform is calculated as: 
\begin{align}
F(\omega_k) \!=\! \frac{1}{{N}} \!\!\sum_{t=0}^{N-1} \! x_te^{-j\omega_k t}\ 
=A(\omega_k) \exp[j\theta(\omega_k)]
\end{align}
where $\omega_k= \frac{2\pi k}{N}$ is the angular frequency, $A(\omega_k)$ is the amplitude spectrum, and $\theta(\omega_k)$ is the phase spectrum.
For perturbations in amplitude spectrum $A(\omega_k)$, the amplitude values of randomly selected segments are replaced with Gaussian noise by considering the original mean and variance in the amplitude spectrum.
While for perturbations in phase spectrum $\theta(\omega_k)$, the phase values of randomly selected segments are added by an extra zero-mean Gaussian noise in the phase spectrum. The amplitude and phase perturbations (APP) based data augmentation combined with aforementioned time-domain augmentation methods bring significant time series anomaly detection improvements as shown in the experiments of~\cite{jingkun20_TAD}.

Another recent work in~\cite{Lee2019} proposes to utilize the surrogate data to improve the classification performance of rehabilitative time series in deep neural network. Two conventional types of surrogate time series are adopted in the work: the amplitude adjusted Fourier transform (AAFT) and the iterated AAFT (IAAFT)~\cite{Schreiber2000}. 
The main idea is to perform random phase shuffle in phase spectrum after Fourier transform and then perform rank-ordering of time series after inverse Fourier transform. The generated time series from AAFT and IAAFT can approximately preserve the temporal correlation, power spectra, and the amplitude distribution of the original time series. 
In the experiments of~\cite{Lee2019}, the authors conducted two types of data augmentation by extending the data by 10 then 100 times through AAFT and IAAFT methods, and demonstrated promising classification accuracy improvements compared to the original time series without data augmentation.




\subsection{Time-Frequency Domain} \label{sec:tf_domain}
Time-frequency analysis is a widely applied technique for time series analysis, which can be utilized as an appropriate input features in deep neural networks. However, similar to data augmentation in frequency domain, only a few studies considered data augmentation from time-frequency domain for time series. 

The authors in~\cite{steven2018feature} adopt short Fourier transform (STFT) to generate time-frequency features for sensor time series, and conduct data augmentation on the time-frequency features for human activity classification by a deep LSTM neural network.
Specifically, two augmentation techniques are proposed. One is the local averaging based on a defined criteria with the generated features appended at the tail end of the feature set. Another is the shuffling of feature vectors to create variation in the data. Similarly, in speech time series, recently SpecAugment~\cite{park2019specaugment} is proposed to make data augmentation in Mel-Frequency (a time-frequency representation based on STFT for speech time series), where the augmentation scheme consists of warping the features, masking blocks of frequency channels, and masking blocks of time steps. They demonstrate that SpecAugment can greatly improve the performance of speech recognition neural networks and obtain state-of-the-art results.

\begin{figure}[!t]
    \centering
    \subfigure[time domain]{\includegraphics[width=0.235\textwidth]{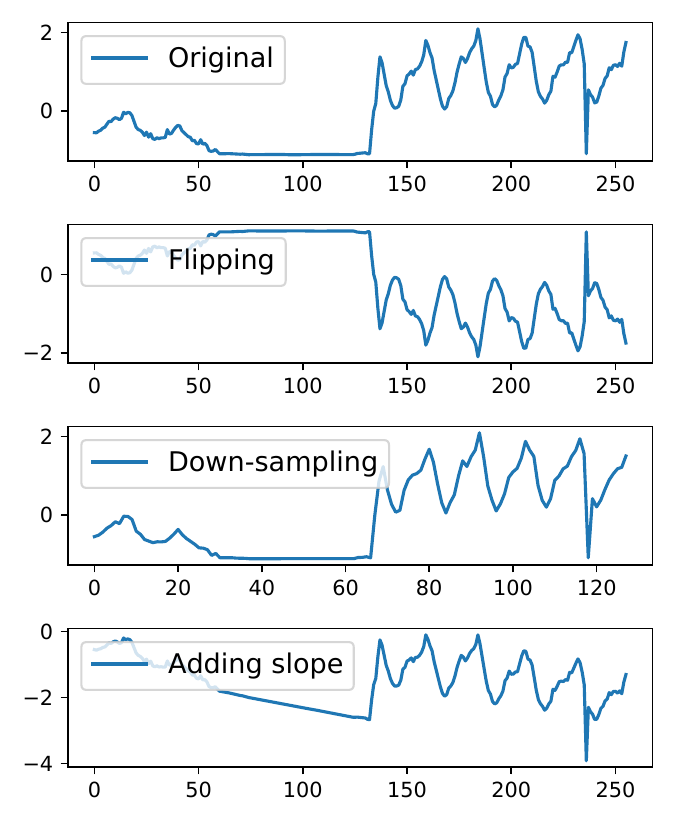}\label{fig:Peri_ACF_oriTS}}
    \subfigure[(time-)frequency domain]{\includegraphics[width=0.235\textwidth]{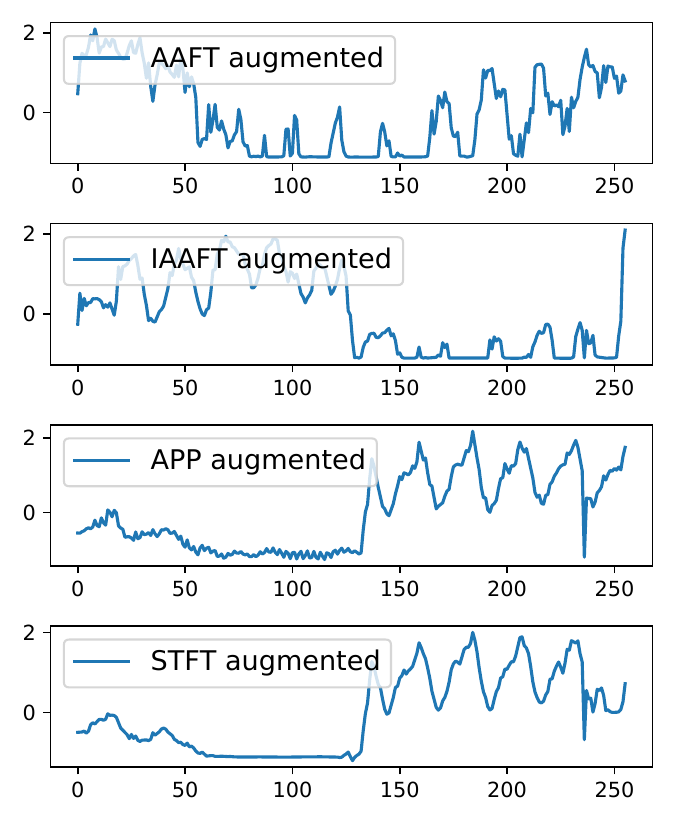}\label{fig:Peri_ACF_Ori}} 
    \vspace{-0.3cm}
    \caption{Illustration of several typical time series data augmentations in time, frequency, and time-frequency domains.}
    \vspace{-0.3cm}
    \label{fig:basic_augmetation}
\end{figure}

For illustration, we summarize several typical time series data augmentation methods in time, frequency, and time-frequency domains in Fig.~\ref{fig:basic_augmetation}.

\vspace{-0.1cm}

\section{Advanced Data Augmentation Methods}
\vspace{-0.1cm}
\subsection{Decomposition-based Methods}
\vspace{-0.05cm}

Decomposition-based time series augmentation has also been adopted and shown success in many time series related tasks, such as forecasting and anomaly detection.
Common decomposition method like STL~\cite{STL_cleveland1990stl} or RobustSTL~\cite{RobustSTL_wen2018robuststl} decomposes time series $x_t$ as
\begin{equation}\label{eq:stl} 
x_t = \tau_t + s_t + r_t, \quad t = 1,2,...N
\end{equation}
where $\tau_t$ is the trend signal, $s_t$ is the seasonal/periodic signal, and the $r_t$ denotes the remainder signal. 

In~\cite{kegel2018feature}, authors discussed the decomposition method to generate new time series. After STL, it recombines new time series with a deterministic component and a stochastic component. The deterministic part is reconstructed by adjusting weights for base, trend, and seasonality. The stochastic part is generated by building a composite statistical model based on residual, such as an auto-regressive model. The summed generated time series is validated by examining whether a feature-based distance to its original signal is within certain range. 
Meanwhile, authors in~\cite{Bergmeir2016} proposed to apply bootstrapping on the STL decomposed residuals to generate augmented signals, which are then added back with trend and seasonality to assemble a new time series. An ensemble of the forecasting models on the augmented time series has outperformed the original forecasting model consistently, demonstrating the effectiveness of decomposition-based time series augmentation approaches.

Recently, in~\cite{jingkun20_TAD}, authors showed that applying time-domain and frequency-domain augmentation on the decomposed residual that is generated using robust decomposition~\cite{FastRobustSTL_wen2020,WenRobustTrend19} can help increase the performance of anomaly detection significantly, compared with the same method without augmentation.




\subsection{Statistical Generative Models}
\vspace{-0.05cm}

Time series augmentation approaches based on statistical generative models typically involve modelling the dynamics of the time series with statistical models. In~\cite{cao2014parsimonious}, authors proposed a parsimonious statistical model, known as mixture of Gaussian trees, for modeling multi-modal minority class time series data to solve the problem of imbalanced classification, which shows advantages compared with existing oversampling approaches that do not exploit time series correlations between neighboring points. Authors in~\cite{smyl2016data} use samples of parameters and forecast paths calculated by a statistical algorithm called LGT (Local and Global Trend). More recently, in~\cite{Kang2019} researchers use mixture autoregressive (MAR) models to simulate sets of time series and investigate the diversity and coverage of the generated time series in a time series feature space. 

Essentially, these models describe the conditional distribution of time series by assuming the value at time $t$ depends on previous points. Once the initial value is perturbed, a new time series sequence could be generated following the conditional distribution.

\subsection{Learning-based Methods}
\vspace{-0.05cm}
Time series data augmentation methods
should be capable of not only generating diverse samples, but also mimicking the characteristics of real data. 
In this section, we summarize some recent learning based schemes that have such potentials. 

\subsubsection{Embedding Space}
In \cite{DeVries2017}, the data augmentation is proposed to perform in the learned embedding space (aka., latent space). It assumes that simple transforms applied to encoded inputs rather than the raw inputs would produce more plausible synthetic data due to the manifold unfolding in feature space. Note that the selection of the representation model in this framework is open and depends on the specific task and data type. When the time series data is addressed, a sequence autoencoder is selected in \cite{DeVries2017}. Specifically, the interpolation and  extrapolation are applied to generate new samples. The first $k$ nearest labels in the transformed space with the same label are identified. Then for each pair of neighboring samples, a new sample is generated which is the linear combination of them. The difference of interpolation and extrapolation lies in the weight selection in sample generation. This technique is particular useful for time series classification as demonstrated in~\cite{DeVries2017}. Recently, another data augmentation method in the embedding space named MODALS (Modality-agnostic Automated Data Augmentation in the Latent Space) is proposed in ~\cite{cheung2021modals}. 
Instead of training an autoencoder to learn the latent space and generate additional synthetic data for training, the MODALS method train a classification model jointly with different compositions of latent space augmentations, which demonstrates superior performance for time series classification problems.






\subsubsection{Deep Generative Models}
Deep generative models (DGMs) have recently been shown to be able to generate near-realistic high-dimensional data objects such as images and sequences. DGMs developed for sequential data, such as audio and text, often can be extended to model time series data. Among DGMs, generative adversarial networks (GANs) are popular methods to generate synthetic samples and increase the training set effectively.
Although the GAN frameworks have received significant attention in many fields, how to generate effective time series data still remains a challenging problem. In this subsection, we briefly review several recent works on GANs for time series data augmentation. 


In \cite{esteban2017real}, a Recurrent GAN (RGAN) and Recurrent Conditional GAN (RCGAN) are proposed to produce realistic real-valued multi-dimensional time series data. The RGAN adopts RNN in the generator and discriminator, while the RCGAN adopts both RNNs conditioned on auxiliary information. Besides desirable performance of RGAN and RCGAN for time series data augmentation, differential privacy can be used in training the RCGAN for stricter privacy guarantees like medicine or other sensitive domains. 
Recently, \cite{NIPS2019Yoon_tsGAN} proposed TimeGAN, a natural framework for generating realistic time series data in various domains. TimeGAN is a generative time series model, trained adversarially and jointly via a learned embedding space with both supervised and unsupervised losses. Specifically, a stepwise supervised loss is introduced to learn the stepwise conditional distributions in data. It also introduces an embedding network to provide a reversible mapping between features and latent representations to reduce the high-dimensionality of the adversarial learning space. Note that the supervised loss is minimized by jointly training both the embedding and generator networks.


\subsubsection{Automated Data Augmentation}

The idea of automated data augmentation is to automatically search for optimal data augmentation policies through reinforcement learning, meta learning, or evolutionary search
~\cite{ratner2017learning,Cubuk_2019_CVPR,zhang2020adversarial,cheung2021modals}. 
The TANDA (Transformation Adversarial Networks for Data Augmentations) scheme in~\cite{ratner2017learning} is designed to
train a generative sequence model over specified transformation functions using reinforcement learning
in a GAN-like framework to generate realistic transformed data points, which yields strong gains over common heuristic data augmentation methods for a range of applications including image recognition and natural language understanding tasks.
~\cite{Cubuk_2019_CVPR} proposes a procedure called AutoAugment to automatically search for improved data augmentation policies in a reinforcement learning framework. It adopts a controller RNN network to predicts an augmentation policy from the search space and another network is trained to achieve convergence accuracy. Then, the accuracy is used as reward to update the RNN controller for better policies in the next iteration.
The experimental results show that AutoAugment improves the accuracy of modern image classifiers significantly in a wide range of datasets.

For time series data augmentation, the MODALS~\cite{cheung2021modals} is designed to find the optimal composition of latent space transformations for data augmentation using evolution search strategy based on population based augmentation (PBA)~\cite{ho2019population}, which demonstrates superior performance on classification problems in continuous and discrete time series data. Another recent work on automated data augmentation is proposed in ~\cite{fons2021adaptive}, where two sample-adaptive automatic weighting schemes are designed specifically for time series data: one learns to weight the contribution of the augmented samples to the loss, and the other selects a subset of transformations based on the ranking of the predicted training loss. Both adaptive policies demonstrate improvement on classification problems in multiple time series datasets.

\vspace{-0.1cm}
\section{Preliminary Evaluation} \label{sec:evaluation}
\vspace{-0.1cm}
In this section, 
we demonstrate preliminary evaluations in three common time series tasks to show the effectiveness of data augmentation for performance improvement.

\vspace{-0.1cm}
\subsection{Time Series Classification}



In this experiment, we compare the classification performance with and without data augmentation. Specifically, we collect $5000$ time series of one-week long and 5-min interval samples with binary class labels (seasonal or non-seasonal) from Alibaba Cloud monitoring system. The data is randomly splitted into training and test sets where training contains $80\%$ of total samples. We train a fully convolutional network~\cite{wang2017time} to classify each time series in the training set. In our experiment, we inject different types of outliers, including spike, step, and slope, into the test set to evaluate the robustness of the trained classifier. The data augmentations methods applied include cropping, warping, and flipping. Table~\ref{tab:aug-in-classification} summarizes the accuracies with and without data augmentation when different types of outliers are injected into the test set. It can be observed that data augmentation leads to $0.1\% \sim 1.9\%$ accuracy improvement.

\begin{table}[!h]
\centering
\footnotesize
\begin{tabular}{c|cccc}
\hline
Outlier injection  & w/o aug & w/ aug &  Improvement \\
\hline
spike     & 96.26\%  & 96.37\% & 0.11\%   \\
step      & 93.70\%  & 95.62\%  & 1.92\%   \\
slope     & 95.84\%  & 96.16\%  & 0.32\%  \\
\hline
\end{tabular}
\vspace{-0.2cm}
\caption{Accuracy improvement from data augmentation under outlier injection in time series classification.}
\vspace{-0.3cm}
\label{tab:aug-in-classification}
\end{table}

\subsection{Time Series Anomaly Detection}

Given the challenges of both \emph{data scarcity} and \emph{data imbalance} in time series anomaly detection, it is beneficial by adopting data augmentation to generate more labeled data. 
We briefly summarize the results in~\cite{jingkun20_TAD}, where a U-Net based network is designed and evaluated on public Yahoo! dataset~\cite{Laptev2015} for time series anomaly detection. The performance comparison under different settings are summarized in Table~\ref{tab:5-ad-table}, including applying the model on the raw data (U-Net-Raw), on the decomposed residuals (U-Net-DeW), and on the residuals with data augmentation (U-Net-DeWA). 
The applied data augmentation methods include flipping, cropping, label expansion, and APP based augmentation in frequency domain. It can be observed that the decomposition helps the increase of the F1 score and the data augmentation further boosts the performance.

\begin{table}[!h]
    \centering
    \footnotesize
\begin{tabular}{c|ccc}
\hline
{Algorithm} &  Precision &  Recall &     F1 \\
\hline
U-Net-Raw &      0.473 &   0.351 &  0.403   \\
U-Net-DeW  &      0.793 &   0.569 &  0.662 \\
U-Net-DeWA (w/ aug) &      0.859 &   0.581 &  0.693  \\
\hline
\end{tabular}
\vspace{-0.1cm}
        \caption{Time series anomaly detection improvement from data augmentation based on precision, recall, and F1 score.}\label{tab:5-ad-table}
        \vspace{-0.3cm}
\end{table}

\vspace{-0.1cm}
\subsection{Time Series Forecasting}
In this subsection we demonstrate the practical effectiveness of data augmentation in two popular deep models DeepAR \cite{salinas2019deepar} and Transformer \cite{vaswani2017attention}. In Table~\ref{tab:aug-in-forecasting}, we report the performance improvement on mean absolute scaled error (MASE) on several public datasets: {\it electricity} and {\it traffic} from UCI Learning Repository\footnote{\url{http://archive.ics.uci.edu/ml/datasets.php}}
and 3 datasets from the M4 competition\footnote{\url{https://github.com/Mcompetitions/M4-methods/tree/master/Dataset}}. We consider the basic augmentation methods including cropping, warping, flipping, and APP based augmentation in frequency domain. 
In Table \ref{tab:aug-in-forecasting}, we summarize average MASE without augmentation, with augmentation and average relative improvement (ARI) which is computed as the mean of $(\textrm{MASE}_{\textrm{w/o aug}} -\textrm{MASE}_{\textrm{w aug}})/\textrm{MASE}_{\textrm{w aug}}$. 
We observe that the data augmentation methods bring promising results for all models in average sense. However, the negative results can still be observed for specific data/model pairs. As a future work, it motivates us to search for advanced automated data augmentation policies that stabilize the influence of data augmentation specifically for time series forecasting.

\begin{table}[!h]
\centering
\scriptsize
\begin{tabular}{l|rrr|rrr}
\hline
\multirow{2}{*}{Dataset}  & \multicolumn{3}{c|}{DeepAR}  & \multicolumn{3}{c}{Transformer} \\
 \cline{2-7}
 &w/o aug &w/ aug &ARI &w/o aug &w/ aug &ARI\\
 \hline
 electricity&$0.87$&$0.97$&$1.92\%$  &$1.04$&$1.11$&$-2\%$    \\
traffic     & $0.66$&$0.80$&$-12\%$   & $0.70$&$0.91$&$-16\%$   \\
m4-hourly   &$6.33$&$5.35$&$56\%$    &$7.77$&$7.87$&$38\%$    \\
m4-daily    &$4.88$&$4.48$&$10\%$   &$7.85$&$7.38$&$37\%$ \\
m4-weekly   &$12.00$&$9.34$&$76\%$    &$6.62$&$7.09$&$23\%$  \\
\hline
\end{tabular}
\vspace{-0.2cm}
\caption{Time seires forecasting improvement from data augmentation based on MASE. }
\vspace{-0.3cm}
\label{tab:aug-in-forecasting}
\end{table}

\vspace{-0.1cm}
\section{Discussion for Future Opportunities}
\subsection{Augmentation in Time-Frequency Domain}

As discussed in Section~\ref{sec:tf_domain}, so far there are only limited studies of time series data augmentation methods based on STFT in the time-frequency domain. Besides STFT, wavelet transform and its variants including continuous wavelet transform (CWT) and discrete wavelet transform (DWT), are another family of adaptive time–frequency domain analysis methods to characterize time-varying properties of time series. Compared to STFT, they can handle non-stationary time series and non-Gaussian noises more effectively and robustly. 
Among many wavelet transform variants, maximum overlap discrete wavelet transform (MODWT) is especially attractive for time series analysis~\cite{percival2000wavelet,WenRobustPeriod20} due to the following advantages: 1) more computationally efficiency compared to CWT; 2) ability to handle any time series length; 3) increased resolution at coarser scales compared with DWT. MODWT based surrogate time series have been proposed in~\cite{PhysRevE06_WIAAFT}, where wavelet iterative amplitude adjusted Fourier transform (WIAAFT) is designed by combining the iterative amplitude adjusted Fourier transform (IAAFT) scheme to each level of MODWT coefficients. In contrast to IAAFT, WIAAFT does not assume sationarity and can roughly maintain the shape of the original data in terms of the temporal evolution. Besides WIAAFT, we can also consider the perturbation of both amplitude spectrum and phase spectrum as~\cite{jingkun20_TAD} at each level of MODWT coefficients as a data augmentation scheme. 

It would be an interesting future direction to investigate how to exploit different wavelet transforms (CWT, DWT, MODWT, etc.) for an effective time-frequency domain based time series data augmentation in deep neural networks.

\subsection{Augmentation for Imbalanced Class}
\vspace{-0.05cm}
In time series classification, class imbalance occurs very frequently.  
One classical approach addressing imbalanced classification problem is to oversample the minority class as the synthetic minority oversampling technique (SMOTE)~\cite{fernandez2018smote} to artificially mitigate the imbalance. However, this oversampling strategy may change the distribution of raw data and cause overfitting. Another approach is to design cost-sensitive model by using adjust loss function~\cite{geng2018cost}. Furthermore, \cite{jingkun20_TAD} designed label-based weight and value-based weight in the loss function in convolution neural networks, which considers weight adjustment for class labels and the neighborhood of each sample. Thus, both class imbalance and temporal dependency are explicitly considered. 

Performing data augmentation and weighting for imbalanced class together would be an interesting and effective direction. A recent study investigates this topic in the area of CV and NLP~\cite{hu2019learning}, which significantly improves text and image classification in low data regime and imbalanced class problems.
In future, it is interesting to design deep network by jointly considering data augmentation and weighting for imbalanced class in time series data.








\subsection{Augmentation Selection and Combination}
\vspace{-0.05cm}

Given different data augmentation methods summarized in Fig.~\ref{fig:taxonomy}, one key strategy is how to select and combine various augmentation methods together. The experiments in \cite{um2017data} show that the combination of three basic time-domain methods (permutation, rotation, and time warping) is better than that of a single method and achieves the best performance in time series classification. Also, the results in~\cite{rashid2019times} demonstrate substantial performance improvement for a time series classification task when using a deep neural network by combining four data augmentation methods (i.e, jittering, scaling, rotation and time-warping). However, considering various data augmentation methods, directly combining different augmentations may result in a huge amount of data, and may not be efficient and effective for performance improvement. 
Recently, RandAugment~\cite{cubuk2019randaugment} is proposed as a practical way for augmentation combination in image classification and object detection. For each random generated dataset, RandAugment is based on only two interpretable hyperparameters $N$ (number of augmentation methods to combine) and $M$ (magnitude for all augmentation methods), where each augmentation is randomly selected from $K$=14 available augmentation methods. Furthermore, this randomly combined augmentation with simple grid search can be used in the reinforcement learning based data augmentation as ~\cite{Cubuk_2019_CVPR} for efficient space searching. 

An interesting future direction is how to design effective augmentation selection and/or combination strategies suitable for time series data in deep learning. Customized reinforcement learning and meta learning optimized for time series could be potential approaches. Furthermore, algorithm efficiency is another important consideration in practice. 


\subsection{Augmentation with Gaussian Processes}
\vspace{-0.05cm}
Gaussian Processes (GPs) \cite{gp} are well-known Bayesian non-parametric models suitable for time series analysis \cite{roberts2013gaussian}. From the function-space view, GPs induce a distribution over functions, i.e., a stochastic process. Time series can be viewed as functions with time as input and observation as output, and thus can be modeled with GPs. A GP $f(t) \sim \mathcal{GP}(m(t), k(t, t'))$ is characterized by a mean function $m(t)$ and a covariance kernel function $k(t, t')$. The choice of the kernel allows to place assumptions on some general properties of the modeled functions, such as smoothness, scale, periodicity and noise level. Kernels can be composed through addition and multiplication, resulting in compositional function properties, such as pseudo-periodicity, additive decomposability, and change point. GPs are often applied to interpolation and extrapolation tasks, which correspond to imputation and forecasting in time series analysis. 
Furthermore, deep Gaussian processes(DGPs)~\cite{damianou2013deep,salimbeni2017doubly}, which are richer models with hierarchical composition of GPs and often exceed standard (single-layer) GPs significantly in many cases, have not been well studied for time series. 
We believe GPs and DGPs are future directions as they allow to sample time series with those properties mentioned above through the design of kernels, and to generate new data instances from existing ones by exploiting their interpolation/extrapolation abilities.

\vspace{-0.1cm}
\subsection{Augmentation with Deep Generative Models}
\vspace{-0.05cm}
Current DGMs adopted for time series data augmentation are mainly GANs. However, other DGMs also have great potentials for time series modeling. For example, deep autoregressive networks (DARNs) exhibit a natural fit for time series because they generate data in a sequential manner, obeying the causal direction of physical time series data generating process. DARNs like Wavenet~\cite{wavenet} and Transformer~\cite{vaswani2017attention} have demonstrated promising performance in time series forecasting tasks~\cite{gluonts}. Another example is normalizing flows (NFs)~\cite{kobyzev2020normalizing}, which recently have shown success in modeling time series stochastic processes with excellent inter-/extrapolation performance given observed data~\cite{deng2020modeling}. Most recently, variational autoencoders (VAEs) based data augmentation~\cite{fu2020data} are investigated for human activity recognition.

In summary, besides the common GAN architectures, how to leverage other deep generative models like DARNs, NFs, and VAEs, which are less investigated for time series data augmentation, remain exciting future opportunities.

\vspace{-0.25cm}
\section{Conclusion}
\vspace{-0.1cm}
As deep learning models are becoming more popular on time series data, the limited labeled data calls for effective data augmentation methods. In this paper, we give a comprehensive survey on time series data augmentation methods in various tasks. We organize the reviewed methods in a taxonomy consisting of basic and advanced approaches, summarize representative methods in each category, compare them empirically in typical tasks, and highlight future research directions.




\bibliographystyle{named}



{\small
\bibliography{7_reference}}



\end{document}